%% file: main.tex
\documentclass[acmtog]{acmart}
\usepackage{booktabs}
\usepackage{tabularx}
\usepackage{xcolor} 
\usepackage{float}
\usepackage{tikz}
\usepackage{xcolor}
\usepackage{graphicx}
\usepackage{booktabs}
\usepackage{colortbl}
\usepackage{caption}
\usepackage{subcaption}
\usepackage{multirow}
\usepackage{booktabs}
\usepackage{microtype}
\usepackage{xcolor} 
\usepackage[most]{tcolorbox}
\usepackage{soul}
\usepackage{enumitem}
\sethlcolor{yellow!25}

\newcommand{\tocite}[1]{{\textcolor{red}{[TOCITE:]}}}

\usepackage{color}

\definecolor{colorfirst}{rgb}{.866,.945, 0.831} %
\definecolor{colorsecond}{rgb}{1, 0.98, 0.83} %
\definecolor{colorthird}{rgb}{0.76, 0.87, 0.92} %

\newcommand{\cellfirst}{\cellcolor{colorfirst}}
\newcommand{\cellsecond}{\cellcolor{colorsecond}}

\usepackage[ruled]{algorithm2e}

\AtBeginDocument{%
  \providecommand\BibTeX{{%
    \normalfont B\kern-0.5em{\scshape i\kern-0.25em b}\kern-0.8em\TeX}}}

\acmSubmissionID{1986}

\setcopyright{acmlicensed}
\begin{document}
\setcopyright{acmlicensed}
\acmJournal{TOG}
\acmYear{2025} \acmVolume{44} \acmNumber{6} \acmArticle{215} \acmMonth{12}\acmDOI{10.1145/3763343}

\title{Split4D: Decomposed 4D Scene Reconstruction Without Video Segmentation}
\author{Yongzhen Hu}
\email{huyongzhen@zju.edu.cn}
\affiliation{
  \institution{Zhejiang University}
  \country{China}
}
\affiliation{
  \institution{Ant Group}
  \country{China}
}

\author{Yihui Yang}
\email{yihuiyang@zju.edu.cn}
\affiliation{
  \institution{Zhejiang University}
  \country{China}
}

\author{Haotong Lin}
\email{haotongl@zju.edu.cn}
\affiliation{
  \institution{Zhejiang University}
  \country{China}
}

\author{Yifan Wang}
\email{accwyf@gmail.com}
\affiliation{
  \institution{Zhejiang University}
  \country{China}
}

\author{Junting Dong}
\email{jtdong@zju.edu.cn}
\affiliation{
  \institution{Shanghai AI Lab}
  \country{China}
}

\author{Yifu Deng}
\email{dengyifu.dyf@antgroup.com}
\affiliation{
  \institution{Ant Group}
  \country{China}
}

\author{Xinyu Zhu}
\email{carina.zxy@antgroup.com}
\affiliation{
  \institution{Ant Group}
  \country{China}
}

\author{Fan Jia}
\email{jiafan.jf@antgroup.com}
\affiliation{
  \institution{Ant Group}
  \country{China}
}

\author{Hujun Bao}
\email{bao@cad.zju.edu.cn}
\affiliation{
  \institution{State Key Lab of CAD\&CG, Zhejiang University}
  \country{China}
}

\author{Xiaowei Zhou}
\email{xwzhou@zju.edu.cn}
\affiliation{
  \institution{State Key Lab of CAD\&CG, Zhejiang University}
  \country{China}
}

\author{Sida Peng}
\authornote{Corresponding author: Sida Peng.}
\email{pengsida@zju.edu.cn}
\affiliation{
  \institution{Zhejiang University}
  \country{China}
}

\begin{teaserfigure}
  \includegraphics[width=\textwidth]{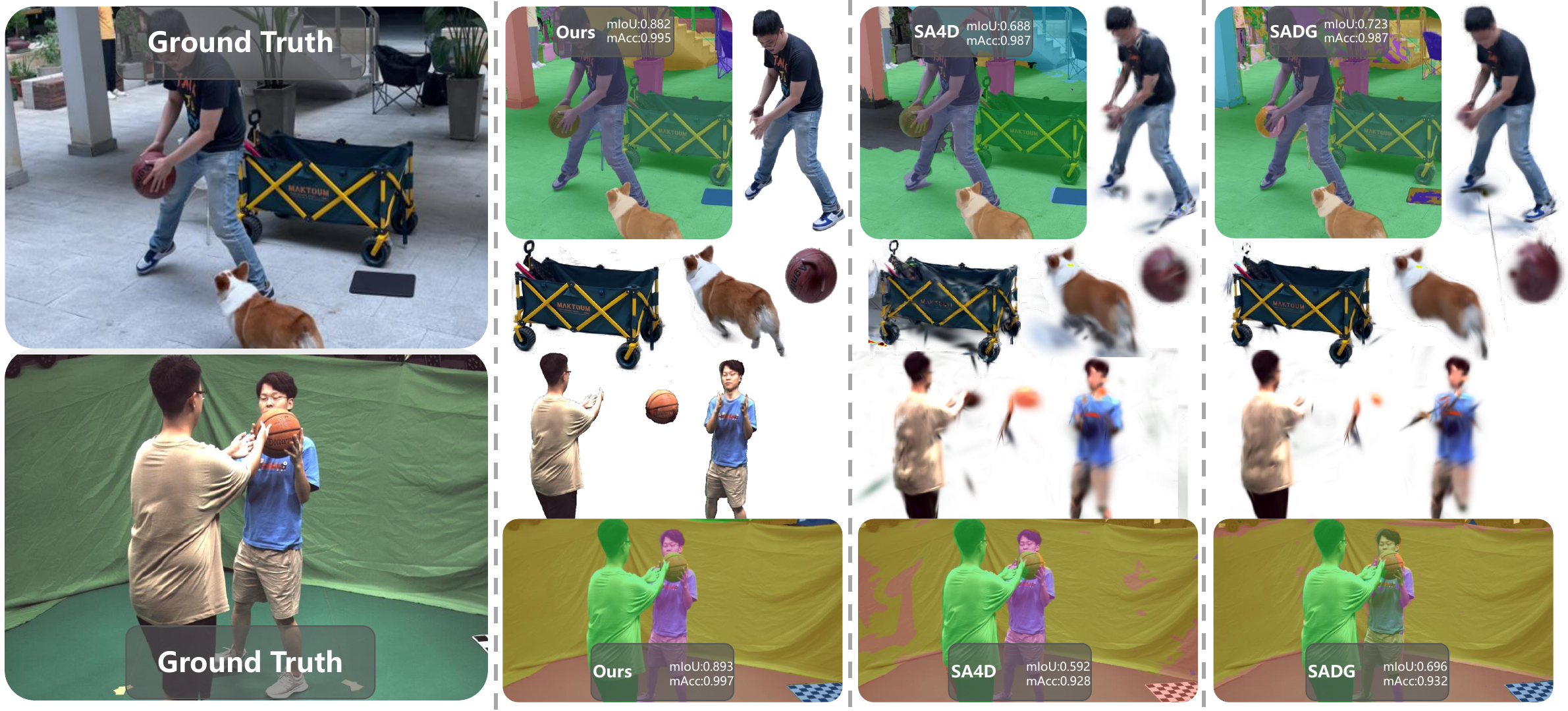}
  \caption{\textbf{Decomposed 4D Scene Reconstruction.} Our method achieves high-quality 4D reconstruction and segmentation on challenging dynamic 3D scenes, particularly in scenarios with multi-human interaction and complex motion. We extract key scene components, encompassing both static and dynamic elements, to support the visual comparison. Compared to recent methods SA4D~\cite{sa4d} and SADG~\cite{sadg}, our approach produces more accurate spatial boundaries and enables higher-fidelity decomposition of complex scenes.}
  \label{fig:teaser}
\end{teaserfigure}

\input{sec/00_abstract}



\maketitle

\input{sec/01_introduction}
\input{sec/02_related_work}
\input{sec/03_method}
\input{sec/04_experiment}

\input{sec/05_application}
\input{sec/06_conclusion}
\input{sec/07_acknowledgments}
\bibliographystyle{ACM-Reference-Format}
\bibliography{reference}

\appendix

\end{document}

%% file: sec/00_abstract.tex
\begin{abstract}

This paper addresses the problem of decomposed 4D scene reconstruction from multi-view videos.
Recent methods achieve this by lifting video segmentation results to a 4D representation through differentiable rendering techniques.
Therefore, they heavily rely on the quality of video segmentation maps, which are often unstable, leading to unreliable reconstruction results.
To overcome this challenge, our key idea is to represent the decomposed 4D scene with the Freetime FeatureGS and design a streaming feature learning strategy to accurately recover it from per-image segmentation maps, eliminating the need for video segmentation.
Freetime FeatureGS models the dynamic scene as a set of Gaussian primitives with learnable features and linear motion ability, allowing them to move to neighboring regions over time.
We apply a contrastive loss to Freetime FeatureGS, forcing primitive features to be close or far apart based on whether their projections belong to the same instance in the 2D segmentation map.
As our Gaussian primitives can move across time, it naturally extends the feature learning to the temporal dimension, achieving 4D segmentation.
Furthermore, we sample observations for training in a temporally ordered manner, enabling the streaming propagation of features over time and effectively avoiding local minima during the optimization process.
Experimental results on several datasets show that the reconstruction quality of our method outperforms recent methods by a large margin.
\end{abstract}

%% file: sec/01_introduction.tex
\section{Introduction}
This paper aims to reconstruct decomposed 4D representations from multi-view videos, enabling precise 4D segmentation of dynamic 3D scenes across space and time. 
Such 4D representation is fundamental for understanding complex real-world environments and supports a wide range of downstream applications, including content creation, robotics, and autonomous driving. 
For instance, decomposed 4D representations allow for selectively removing or translating dynamic objects within a 3D scene, facilitating visual effects that are critical for immersive content creation.

Recently, some methods~\cite{sa4d} first reconstruct 4D representations~\cite{4dgs_huake} from input videos, and then map 2D video segmentation labels into the 4D space through differentiable rendering.
While these approaches have achieved reasonable 4D segmentation quality, they heavily rely on the quality of the input video segmentation. 
In particular, video tracking methods~\cite{DEVA} often struggle to produce long-term consistent segmentations, 
and errors in 2D labels are directly propagated to the 4D representation. Moreover, for multi-view videos inputs, 
ensuring label consistency across different views is non-trivial, making it difficult to obtain coherent 4D segmentation.

Inconsistent segmentation labels are also a challenge in the task of reconstructing decomposed 3D representations from multi-view images.
To address this problem, some methods~\cite{contrastive_lift,omniseg3d} define a 3D feature field and learn it from multi-view segmentation maps with a contrastive loss.
Specifically, for any pair of 3D points, they enforce their feature vectors to be close or far apart according to whether they belong to the same mask in the 2D segmentation maps.
As the point features are independent of specific instance labels, there is no requirement for multi-view consistent segmentation labels.
However, such a feature learning strategy is not easily extensible to 4D scenes, because for any two 3D points at any time, it is not straightforward to determine whether they belong to the same instance.
A solution is to leverage deformable Gaussian splatting~\cite{4dgs_huake, deformable_3dgs} to establish temporal correspondences of 3D points, as presented in SADG~\cite{sadg}.
However, deformable Gaussian splatting struggles to reconstruct dynamic 3D scenes with large motions at high quality.

In this paper, we propose a novel framework, named Split4D, for recovering decomposed 4D representations of complex dynamic scenes from multi-view videos.
Our key idea is to represent the decomposed 4D scene as Freetime FeatureGS and design a streaming feature learning strategy to accurately recover it from segmentation maps.
Specifically, inspired by FreeTimeGS~\cite{ftgs}, we model the dynamic scene as a set of Gaussian primitives with linear motion ability, which can exist at arbitrary spatiotemporal locations and be reused over time via linear movement.
In addition, each Gaussian primitive is assigned a feature vector as its instance label.
These primitive features are trained using a contrastive loss, which enforces any two feature vectors to be close or far apart according to whether their projections belong to the same mask in the 2D segmentation map at a specific viewpoint and time step.
As Gaussian primitives can translate over time, their features will compute contrastive loss with other features at nearby time steps, thus extending the feature learning to 4D. 
This means that point features belonging to the same instance will be close to each other over time, naturally achieving 4D segmentation.
Compared with previous methods~\cite{sa4d}, our method does not require temporally consistent segmentation labels, because our features are not bound to specific instance labels.
In addition, our Freetime FeatureGS can reconstruct and segment complex dynamic 3D scenes faithfully.

Given multi-view videos with segmentation maps, we find that randomly sampling observations for training leads to significant differences in features of the same object at time steps with large temporal gaps.
A plausible explanation is that Freetime FeatureGS only has short-range correspondences, and features at time steps with large temporal gaps lack associations.
If we randomly sample a time step for training, the features of the same instance at different time steps probably converge to different points, leading to inaccurate 4D segmentation.
To overcome this problem, we propose a streaming feature learning strategy, which samples observations for training in a temporally ordered manner, enabling the streaming propagation of features across time and helping avoid local minimum during the optimization process.
The proposed training strategy effectively achieves temporally coherent segmentation, as demonstrated by our experimental results.

To validate the effectiveness of our method, we conducted 4D segmentation experiments on the Neural3DV~\cite{neural3dv}, Multi-Human~\cite{multi_human} and \textit{SelfCap}~\cite{ftgs} datasets.  
The experimental results demonstrate that our proposed novel 4D decomposed representation achieves state-of-the-art performance.

%% file: sec/02_related_work.tex
\section{Related Work}

\subsection{Dynamic Scene 4D Representation}
Most prior work on dynamic scene reconstruction has focused on building holistic 4D representations that capture the overall geometry and appearance of dynamic environments. 
Early approaches, such as those using RGB-D cameras and non-rigid tracking~\cite{dynamicfusion, Holoportation, fusion4d}, build TSDF models in canonical space but were limited in robustness and scalability. 
With the advent of neural implicit representations~\cite{nerf}, 
subsequent NeRF variants have extended to dynamic scenes.
Some methods~\cite{d_nerf, nerfies, hypernerf, neural3dv} learn deformation fields that warp a canonical radiance field to match frame-wise observations,
while others~\cite{neural3dv, K-planes, Hexplane} embed time conditioning to model dynamic scenes directly.
Another line is to model specific objects, such as dynamic human models by introducing body priors~\cite{neural_body}.
However, these methods primarily focus on unified scene representations, and still face challenges in rendering quality and computational efficiency.
Recently, Gaussian-based~\cite{3dgs} representations have attracted increasing attention for their efficiency and high rendering quality. Some methods~\cite{4dgs_huake, deformable_3dgs, per_gaussian_embedding, motion_aware_3dgs, 3d_geometry_aware} extend 3D Gaussian Splatting to dynamic scenes by introducing global deformation in canonical space, 
while others~\cite{fudan_4dgs, space_time_gaussian, ftgs} propose more flexible 4D Gaussian primitives for dynamic reconstruction. 
Nevertheless, these approaches rarely tackle the need for decomposed, instance-aware 4D representations. In contrast, our work is dedicated to learning a flexible decomposed 4D feature field, which enables accurate and temporally consistent instance segmentation as well as efficient scene editing in dynamic environments.

\subsection{3D Segmentation in Static Scenes}
3D segmentation has made significant progress in recent years, with methods developed for various 3D representations such as RGB-D images~\cite{depth_aware_cnn, malleable_2_5d_convolution}, voxels~\cite{occuseg,point_voxel_cnn}, and point clouds~\cite{randle_net, object_bounding_boxes}. 
However, the scarcity of large-scale annotated 3D datasets~\cite{Scannet} limits the diversity and generalizability of these models, confining most approaches to closed-set segmentation within static scenes. 
To address data limitations, recent works have explored lifting 2D segmentation results from powerful foundation models~\cite{segment_anything, CLIP, Dinov2, maskformer, dino}, into 3D space for scene understanding, typically via differentiable rendering.
Some methods~\cite{semantic_nerf, Lerf, feature_3dgs, open_nerf, langsurf} focus on semantic segmentation by distilling 2D features~\cite{CLIP, Dinov2, dino} into 3D radiance or feature fields, leveraging architectures like NeRF or Gaussian Splatting.
Others aim for instance-level segmentation, but often struggle with label consistency across views. Recent works address this by preprocessing~\cite{gaussian_grouping, cosseggaussians, panoptic_lifting, Gaga} to generate view-consistent labels via video tracker~\cite{DEVA} or matching~\cite{Gaga,Pointmap_matching}.
Alternatively, other studies~\cite{omniseg3d, saga, unified_lift, contrastive_lift, pcf_lift} sidestep explicit cross-view correspondences by first using a contrastive loss to learn a 3D-consistent feature field, and then post-processing it with clustering algorithms, which achieve impressive segmentation accuracy.
However, the majority of these approaches remain restricted to static scenes and do not extend to dynamic environments. Our work aims to bridge this gap by enabling instance-aware 4D segmentation in dynamic scenes.

\subsection{Instance-level 4D Segmentation and Scene Editing}
Instance-level 4D segmentation and editing in dynamic scenes remains underexplored. 
Some methods~\cite{ST-NeRF, multi_human, neuphysics} construct scene graphs by training a separate radiance field for each object, or simply segment the dynamic foreground from the static background. However, these approaches~\cite{ST-NeRF, multi_human} typically rely on strong priors, limiting their generalizability to complex dynamic scenes.
Other methods~\cite{DGD, 4d-editor, 4LEGS} distill features from foundation models to build semantic feature fields, often incur high computational costs which restricts their scalability and practical applicability.
Recently, two concurrent works, SA4D~\cite{sa4d} and SADG~\cite{sadg}, have made progress towards instance-level 4D segmentation. SA4D leverages video tracking~\cite{DEVA} to obtain temporally consistent instance labels and maps 2D segmentations into a learned 4D feature field. 
SADG compresses the 4D scene into a canonical space using deformable 4D Gaussians and employs contrastive learning to construct a 4D-consistent feature field. 
However, SA4D heavily depends on the quality of video tracking, which is often unreliable in long-term segmentation scenarios, while SADG relies on deformable 4D Gaussian representations, which struggle to establish consistent feature associations for objects with complex motions.
In contrast, our method introduces a novel framework, Split4D, which avoids reliance on video trackers and does not require deformable 4D Gaussian representations, enabling accurate and temporally consistent instance segmentation, as well as efficient scene editing in complex dynamic scenes.

\begin{figure*}[t]
    \centering
    \includegraphics[width=\textwidth]{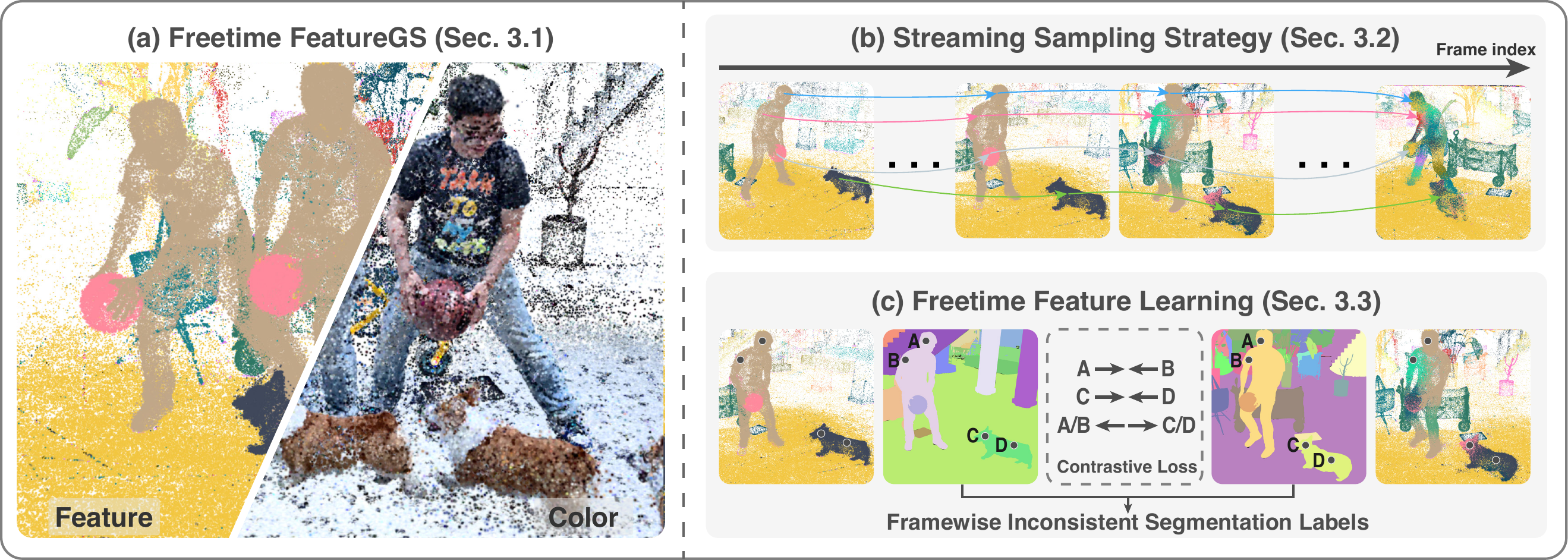}
    \caption{\textbf{Overview of Our Pipeline.} 
    (a) The decomposed 4D representation is constructed based on spatio-temporally free Gaussian primitives, each associated with a learnable feature vector.
    (b) During training, a streaming sampling strategy selects temporally ordered frames, ensuring that features propagate smoothly along their motion trajectories across time.
    (c) Exploiting Freetime FeatureGS's local temporal continuity, we impose a contrastive loss on frame-wise inconsistent 2D segmentation labels, enabling 4D feature learning.
    }
    \label{fig:method_overview}
\end{figure*}

%% file: sec/03_method.tex
\section{Method}

Given a multi-view video with corresponding segmentation maps, our goal is to learn a decomposed 4D representation that enables accurate segmentation of dynamic 3D scenes over time.
The segmentation maps are obtained using the Segment Anything Model (SAM)~\cite{segment_anything}.
An overview of our framework is illustrated in Fig.~\ref{fig:method_overview}.
First, we introduce our novel 4D decomposed representation (Sec.~\ref{sec:freetime_featuregs}) and the streaming sampling method (Sec.~\ref{sec:streaming_sampling}) to obtain a consistent feature field across viewpoints and time based on the feature learning strategy (Sec.~\ref{sec:freetime_feature_learning}).
In Sec.~\ref{sec:tracking_regularization}, we present two regularization strategies to enhance the robustness of the feature field learning process.
In Sec.~\ref{sec:loss_and_filter}, we describe our training objective and a filtering strategy for instance processing.

\subsection{Freetime FeatureGS} \label{sec:freetime_featuregs}

Inspired by FreeTimeGS~\cite{ftgs}, which achieves high-quality reconstruction of dynamic scenes, we propose a novel decomposed 4D representation based on free Gaussian primitives. 
These primitives can appear at any time and location in the 4D space, allowing for a more flexible and adaptive modeling of dynamic content.
Specifically, each Gaussian primitive is defined by a 4D mean $\boldsymbol{\mu} \in \mathbb{R}^4$, a 4D scale $\boldsymbol{s} \in \mathbb{R}^4$, an opacity scalar $\alpha \in \mathbb{R}$, left and right quaternions $\mathbf{q}_l, \mathbf{q}_r \in \mathbb{R}^4$, and spherical harmonic coefficients $\mathbf{h} \in \mathbb{R}^m$, where $m$ denotes the number of SH bases.
To model temporal dynamics, each Gaussian is assigned a velocity vector $\boldsymbol{v} \in \mathbb{R}^3$. Given its initial timestamp $\mu_t$, the spatial position of the Gaussian at any time $t$ is computed as:
\begin{equation}
    \boldsymbol{\mu}_x(t) = \boldsymbol{\mu}_x + \boldsymbol{v} \cdot (t - \mu_t),
\end{equation}
where $\boldsymbol{\mu}_x$ and $\mu_t$ represent the spatial position and time of the Gaussian primitive. This formulation enables each Gaussian to move linearly in 3D space over time.

To achieve instance-aware 4D segmentation, we associate each Gaussian with a learnable feature vector $\mathbf{f} \in \mathbb{R}^d$. 
These features are rendered into 2D feature maps through differentiable rasterization:
\begin{equation}
    F_s = \sum_{i \in \mathcal{N}} \mathbf{f}_i \alpha_i T_i,
    \quad \text{where} \quad T_i = \prod_{j=1}^{i-1} (1 - \alpha_j),
\end{equation}
where $\mathcal{N}$ is the set of sorted Gaussian primitives associated with a target pixel of the 2D feature map.

\subsection{Streaming Sampling Strategy} \label{sec:streaming_sampling}

\begin{figure*}[t!]
    \centering
    \includegraphics[width=\textwidth]{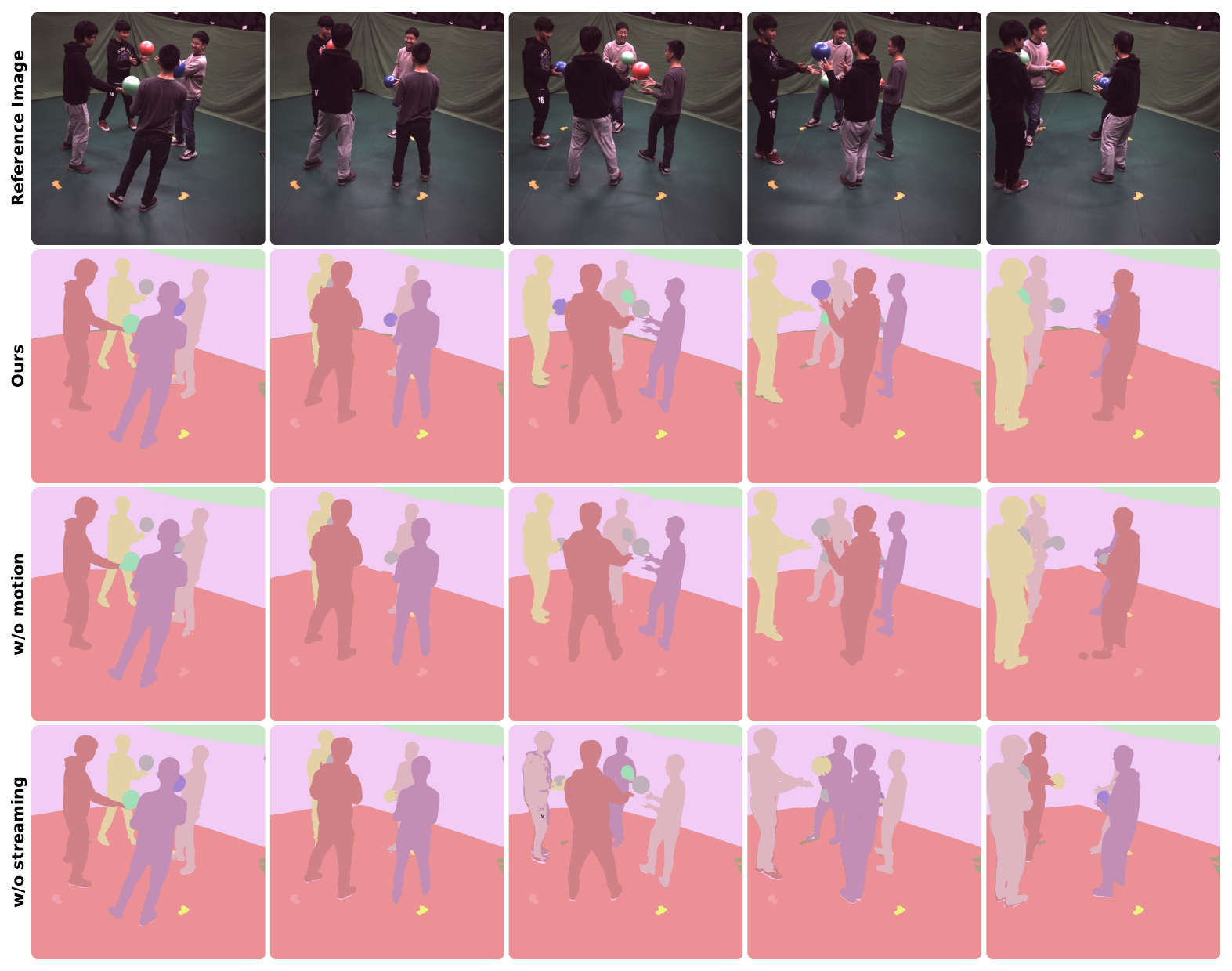}
    \caption{\textbf{Analysis of Motion Modeling and Streaming Learning.} We compare our full model with variants that remove motion modeling and streaming learning strategy. The results demonstrate that both components are crucial for handling complex motions and maintaining temporal consistency in feature learning. The full model achieves the best performance, while removing either component leads to significant performance degradation.}
    \label{fig:ablation_streaming_and_motion}
\end{figure*}

The linear motion and reuse of Gaussian primitives in our 4D representation naturally leads to a gradual replacement process for objects over time. 
This process creates strong local temporal continuity in the feature field, as adjacent frames typically share overlapping primitives. 
However, the local continuity presents a challenge for feature learning: while nearby frames maintain feature consistency through shared primitives, longer temporal intervals may involve complete replacement of primitives representing the same object, as shown in Fig.~\ref{fig:ablation_streaming_and_motion}. 
Direct application of feature learning across such intervals can lead to temporal inconsistency.

To address this challenge, we propose a streaming feature learning strategy that leverages the local continuity to propagate features reliably across time. 
Given feature maps $F_{v,t}$ and corresponding segmentation maps $M_{v,t}$, we construct a temporally ordered sequence of training samples:
\begin{equation}
\mathcal{S} = \{ (F_{v,t}, M_{v,t}) \}_{v=1, t=1}^{V, T},
\end{equation}
where $T$ is the number of frames and $V$ is the number of views.
During training, we sample data pairs from this sequence in a streaming manner to compute the contrastive loss.

\subsection{Freetime Feature Learning} \label{sec:freetime_feature_learning}

For each instance in the segmentation map, we compute its feature center $\bar{\mathbf{f}}_i$ by averaging the features of $N_s$ uniformly sampled points within the corresponding mask. 
The instance-level features are then optimized through the following contrastive clustering loss~\cite{InfoNCE,omniseg3d,contrastiveloss}:
\begin{equation}
\mathcal{L}_{\text{CC}} = - \frac{1}{N_{\text{inst}} N_s} \sum_{i=1}^{N_{\text{inst}}} \sum_{j=1}^{N_s} \log \frac{\exp(\mathbf{f}_{i}^j \cdot \bar{\mathbf{f}}_i / \phi_i)}{\sum_{k=1}^{N_{\text{inst}}} \exp(\mathbf{f}_{i}^j \cdot \bar{\mathbf{f}}_k / \phi_k)},
\end{equation}
where \( N_{\text{inst}} \) is the number of instances, $\mathbf{f}_{i}^j$ is the j-th feature of the i-th instance, and \( \phi_i \) is a temperature parameter computed as:
$\phi_i = \frac{1}{N_s \log(N_s + 10)} \sum_{j=1}^{N_s} \| \mathbf{f}_{i,j} - \bar{\mathbf{f}}_i \|^2$.

Note that although we only require per-image segmentation as input, our model can produce high-quality decomposed 4D reconstruction results.
This is attributed to two factors.
First, for any two 3D points at the same time, the contrastive loss forces their feature vectors to be close or far apart based on whether they belong to the same instance mask in the 2D segmentation map at a specific viewpoint.
Second, as the Gaussian primitives in Freetime FeatureGS can move over time, the contrastive loss computes the distance between features at nearby time steps.
This naturally extends the contrastive loss to the temporal dimension, ensuring that point features belonging to the same instances remain close to each other over time, thus achieving 4D segmentation.

\subsection{Additional Regularization} \label{sec:tracking_regularization}
While our streaming feature learning framework enables robust feature propagation across most dynamic scenes, objects undergoing significant nonlinear motion may break the local temporal continuity of Freetime FeatureGS.
To address these challenges, we introduce two regularization strategies. 
The first strategy focuses on maintaining temporal coherence by aligning features between spatially adjacent Gaussian primitives across frames. 
The second strategy leverages pre-trained semantic features to provide additional guidance during the early stages of training, helping to establish more stable long-range feature associations.

\paragraph{3D Tracking Regularization}
To maintain temporal coherence in the feature field, we propose a 3D tracking-based regularization strategy that aligns features between consecutive frames. 
The key idea is to identify and match Gaussian primitives across adjacent frames based on their spatial proximity and temporal behavior. 
For each frame at time \( t \), we first identify Gaussian primitives that are about to disappear in the next frame \( t+1 \) by computing their temporal opacity: $\alpha(t) = \exp\left( -\frac{1}{2} \left( \frac{t - \mu_t}{s} \right)^2 \right)$,
where \( \mu_t \) is the Gaussian's temporal center and \( s \) its duration. 
This formulation naturally captures the gradual fading of Gaussian primitives over time.

For each disappearing Gaussian at time \( t \), we find its spatial-temporal match in frame \( t+1 \) using k-nearest neighbors search. 
We then enforce feature consistency between these matched pairs through an alignment loss:
\begin{equation}
L_{\text{align}} = \sum \left\| \mathbf{f}_{g_t} - \mathbf{f}_{g_{t'}} \right\|,
\end{equation}
where \( \mathbf{f}_{g_t} \) and \( \mathbf{f}_{g_{t'}} \) are the feature vectors of matched Gaussians.

\paragraph{DINOv2 Feature Alignment Regularization}  
While the 3D tracking regularization helps maintain temporal coherence, we further enhance the feature learning process by incorporating semantic guidance from DINOv2~\cite{Dinov2}. 
To integrate DINOv2 features into our framework, we first project them to match the dimensionality of our feature space \( F_{v,t} \) using a fixed linear projection. 
We then apply a feature learning strategy that encourages consistency between our learned instance features and the projected DINOv2 features. 
For each instance in \( M_{v,t} \), we compute its DINOv2 feature center \( \bar{\mathbf{f}}_{\text{dino},i} \) and optimize the following contrastive loss:
\begin{equation}
\mathcal{L}_{\text{DINO}} = - \frac{1}{N_{\text{inst}} N_s} \sum_{i=1}^{N_{\text{inst}}} \sum_{j=1}^{N_s} \log \frac{\exp(\mathbf{f}_{i}^j \cdot \bar{\mathbf{f}}_{\text{dino},i} / \phi_i)}{\sum_{k=1}^{N_{\text{inst}}} \exp(\mathbf{f}_{i}^j \cdot \bar{\mathbf{f}}_{\text{dino},k} / \phi_k)},
\end{equation}
where $\mathbf{f}_{i}^j$ is the j-th feature of the i-th instance.

This regularization acts as a semantic ``dictionary" that guides our feature space to form a meaningful structure. 
However, since our features may need to deviate from DINOv2's generic representations as training progresses, we only apply this regularization during the early stages of training.

\subsection{Training and Inference} \label{sec:loss_and_filter}
\paragraph{Training}
Our training process focuses on optimizing the instance-level features while keeping the pre-trained 4D reconstruction model fixed. 
This approach allows us to leverage the high-quality geometric reconstruction while learning discriminative features for segmentation. 
The overall training objective combines our contrastive loss with the two regularization terms:
\begin{equation}
\mathcal{L}_{\text{total}} = \mathcal{L}_{\text{CC}} + \lambda_1 \mathcal{L}_{\text{align}} + \lambda_2 \mathcal{L}_{\text{DINO}},
\end{equation}
where \( \lambda_1 \) and \( \lambda_2 \) are loss weights that balance the contribution of each component. The contrastive loss \( \mathcal{L}_{\text{CC}} \) drives the main feature learning process, while \( \mathcal{L}_{\text{align}} \) and \( \mathcal{L}_{\text{DINO}} \) provide temporal coherence and semantic guidance respectively.

\paragraph{Inference}
To achieve accurate 4D segmentation, we develop a dynamic perception filtering strategy that combines feature-based clustering with motion and spatial cues. 
The process begins by randomly sampling a small subset (2–10\%) of Gaussian primitives from the first frame. 
Then, they are clustered using HDBSCAN~\cite{hdbscan}, with each cluster represented by the average feature.

While instance features form the primary basis for segmentation, we observe that incorporating velocity and spatial information can enhance accuracy, particularly for dynamic objects. 
This observation leads us to design a filtering function that considers multiple cues:
\begin{equation}
\text{Filter}(g_i) = \mathbb{I} \left( (\Delta v_i > 3 \sigma_v) \, \land \, (\Delta p_i > 3 \sigma_p) \, \land \, (s(\mathbf{f}_i, \bar{\mathbf{f}}_i) < \tau_{\text{sim}}) \right),
\end{equation}
where \(g_i\) denotes the \(i\)-th Gaussian primitive, \( \Delta v_i \) and \( \Delta p_i \) represent the deviations in velocity and position from the cluster mean, 
\( \sigma_v \) and \( \sigma_p \) are their respective variances within the cluster, \( s(\mathbf{f}_i, \bar{\mathbf{f}}_i) \) measures the feature similarity to the cluster center and
\(\tau_{\text{sim}}\) is a threshold for feature similarity.

This filtering strategy serves as an auxiliary mechanism that complements our feature-based segmentation. 
Rather than using velocity and spatial cues as primary decision factors, we employ them as complementary signals to refine instance-level predictions. 
This approach effectively removes noisy primitives and improves the quality, particularly in challenging scenes where motion and spatial relationships provide valuable additional context.

%% file: sec/04_experiment.tex
\section{Experiment}
\subsection{Datasets and Implementation Details}
\paragraph{Datasets}  
We evaluate our method on three challenging datasets: Neural3DV~\cite{neural3dv}, Multi-Human~\cite{multi_human}, and \textit{SelfCap}~\cite{ftgs}.
\textbf{Neural3DV} comprises six dynamic scenes featuring a subject performing cooking tasks in a kitchen environment, captured by 19–21 synchronized cameras at 30 FPS.  
For our experiments, we utilize the first 300 frames per scene.
\textbf{Multi-Human} contains four complex scenes characterized by intricate human interactions, including large-scale body movements and human-object interactions such as ball passing.  
We process 200 frames per scene at full resolution.
\textit{\textbf{SelfCap}} presents challenging scenarios, such as pet interaction, bicycle repair, and dance performances.
We select three representative scenes, each consisting of 60 frames captured by 22–24 cameras, downsampled to half resolution to maintain consistency with our computational requirements.
Since these datasets were originally designed for 4D reconstruction, they lack ground-truth segmentation labels.  
To enable quantitative evaluation, we employ SAM2 and Grounded-SAM-2 to generate consistent object masks across frames in the test views.  
For each scene, we carefully select 5 to 20 annotated instances for evaluation, encompassing both dynamic and static objects.

\paragraph{Implementation Details}
We implement our method using PyTorch and employ the Adam optimizer for training instance-level features. The learning rate is scheduled from 2.5e\text{-}3 to 2e\text{-}6 to ensure stable convergence.  
We set the instance feature dimension to 32, striking a balance between representation capacity and computational efficiency. All experiments are conducted on an RTX 4090 GPU, with each scene requiring approximately 10{,}000 iterations over 300 frames, taking approximately 30 minutes per scene. 
The loss weights are initialized as \( \lambda_1 = 100 \) and \( \lambda_2 = 500 \) by default.  
For the Multi-Human dataset, we reduce \( \lambda_2 \) to 10 to mitigate the potential interference from DINOv2's semantic priors, which may unintentionally merge distinct individuals into a single semantic category during instance segmentation. \(\tau_{\text{sim}}\) is set to 0.5.

\begin{table*}[t!]
    \centering
    \caption{\textbf{Quantitative comparison on Neural3DV, Multi-Human, and SelfCap datasets.}
    We report mIoU, mAcc, Recall, F$_1$ scores and $\text{Recall}_{\text{dyn}}$ for each method. Green and yellow cell colors indicate the best and second best performance, respectively.}
    \small
    \resizebox{0.97\textwidth}{!}{%
    \begin{tabular}{l|ccccc|ccccc|ccccc}
    \toprule
    \multirow{2}{*}{Metrics} & \multicolumn{5}{c|}{\textbf{Neural3DV~\cite{neural3dv}}} & \multicolumn{5}{c|}{\textbf{Multi-Human~\cite{multi_human}}} & \multicolumn{5}{c}{\textbf{SelfCap~\cite{ftgs}}} \\ 
    \cmidrule(lr){2-6} \cmidrule(lr){7-11} \cmidrule(lr){12-16}
     & Ours  & SA4D  & SADG  & DGD & Omniseg3D
     & Ours  & SA4D  & SADG  & DGD & Omniseg3D
      & Ours  & SA4D  & SADG & DGD & Omniseg3D\\
    \midrule
    mIoU &  \cellfirst \textbf{0.801} & 0.668 & 0.649 & 0.460 & \cellsecond 0.700 & \cellfirst \textbf{0.893}  & 0.592 & \cellsecond 0.696 & 0.379 & 0.533 & \cellfirst \textbf{0.882} & 0.688  & 0.723 & 0.499 & \cellsecond 0.777 \\
    mAcc &  \cellfirst \textbf{0.998} & 0.987 & 0.985 & 0.982 & \cellsecond 0.996 & \cellfirst \textbf{0.997}  & 0.928 & \cellsecond 0.932 & 0.889 & 0.903 & \cellfirst \textbf{0.995} & 0.987  & 0.987 & 0.977 & \cellsecond 0.988\\
    Recall  & \cellsecond 0.903 & 0.861 & 0.827 & 0.773 &  \cellfirst \textbf{0.948} & \cellfirst \textbf{0.948}  & 0.761 & \cellsecond 0.793 & 0.586 & 0.676 & \cellfirst \textbf{0.957} & 0.759  & 0.884 & 0.689 & \cellsecond 0.944\\
    F$_1$  & \cellfirst \textbf{0.870} & 0.759 & 0.744 & 0.595 & \cellsecond 0.807 & \cellfirst \textbf{0.932}  & 0.710 & \cellsecond 0.777 & 0.506 & 0.641 & \cellfirst \textbf{0.936} & 0.797  & 0.827 & 0.648 & \cellsecond 0.869\\
    \midrule
    Recall$_\text{dyn}$ & \cellfirst \textbf{0.833} & 0.812 & \cellsecond 0.813 & 0.627 & 0.797 & \cellfirst \textbf{0.942} & \cellsecond 0.856 & 0.727 & 0.566 & 0.725 & \cellfirst \textbf{0.926} & 0.819 & 0.860 & 0.586 & \cellsecond0.894\\
    \bottomrule
    \end{tabular}
    }
    \label{tab:comparison_result}
    \end{table*}

\begin{figure*}[htbp]
    \centering
    \includegraphics[width=0.93\textwidth]{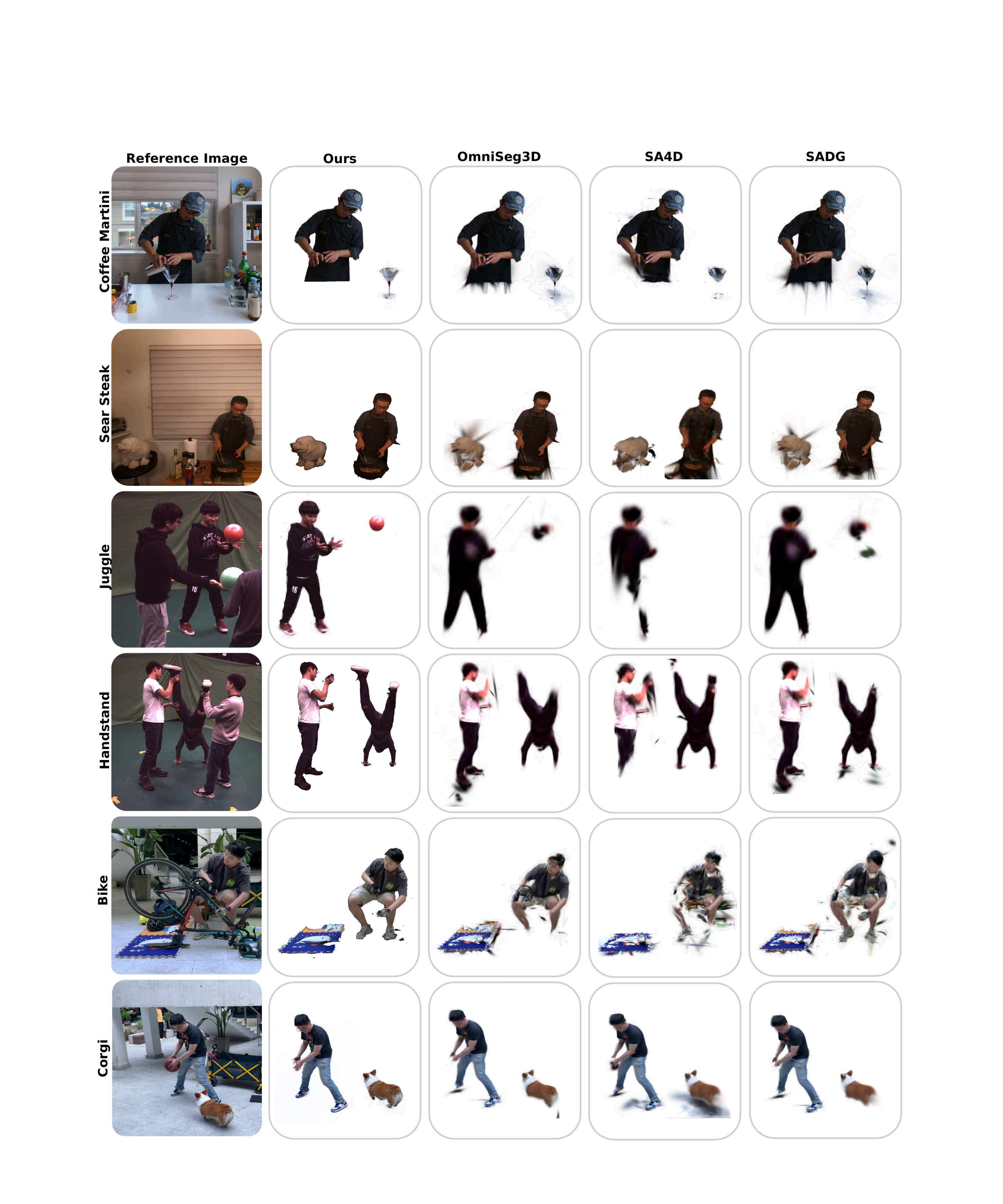}
    \caption{%
        \textbf{Qualitative Comparison Results.}
        We visually compare our method with several baselines, including OmniSeg3D~\cite{omniseg3d}, SA4D~\cite{sa4d} with post-processing, and SADG~\cite{sadg}, as well as with Ground Truth. 
        The results are organized by dataset, with each dataset's results spanning two rows: Neural3DV (top two rows), Multi-Human (middle two rows), and SelfCap (bottom two rows). 
        Our method consistently achieves the best segmentation performance across all datasets, especially on Multi-Human and SelfCap, where scenes involve complex motions.
    }
    \label{fig:comparison_results}
\end{figure*}

\begin{figure}[!tp]
    \centering
    \includegraphics[width=0.48\textwidth]{figures/comp_invert_mask.pdf}
    \caption{
        \textbf{Comparison of Boundary Segmentation in Human-Object Interaction.
        }The circled regions highlight areas of hand-object contact, where our method produces significantly more accurate boundaries.
    }
    \label{fig:ablation_invert_mask}
\end{figure}

\begin{figure}[t!]
    \includegraphics[width=0.47\textwidth]{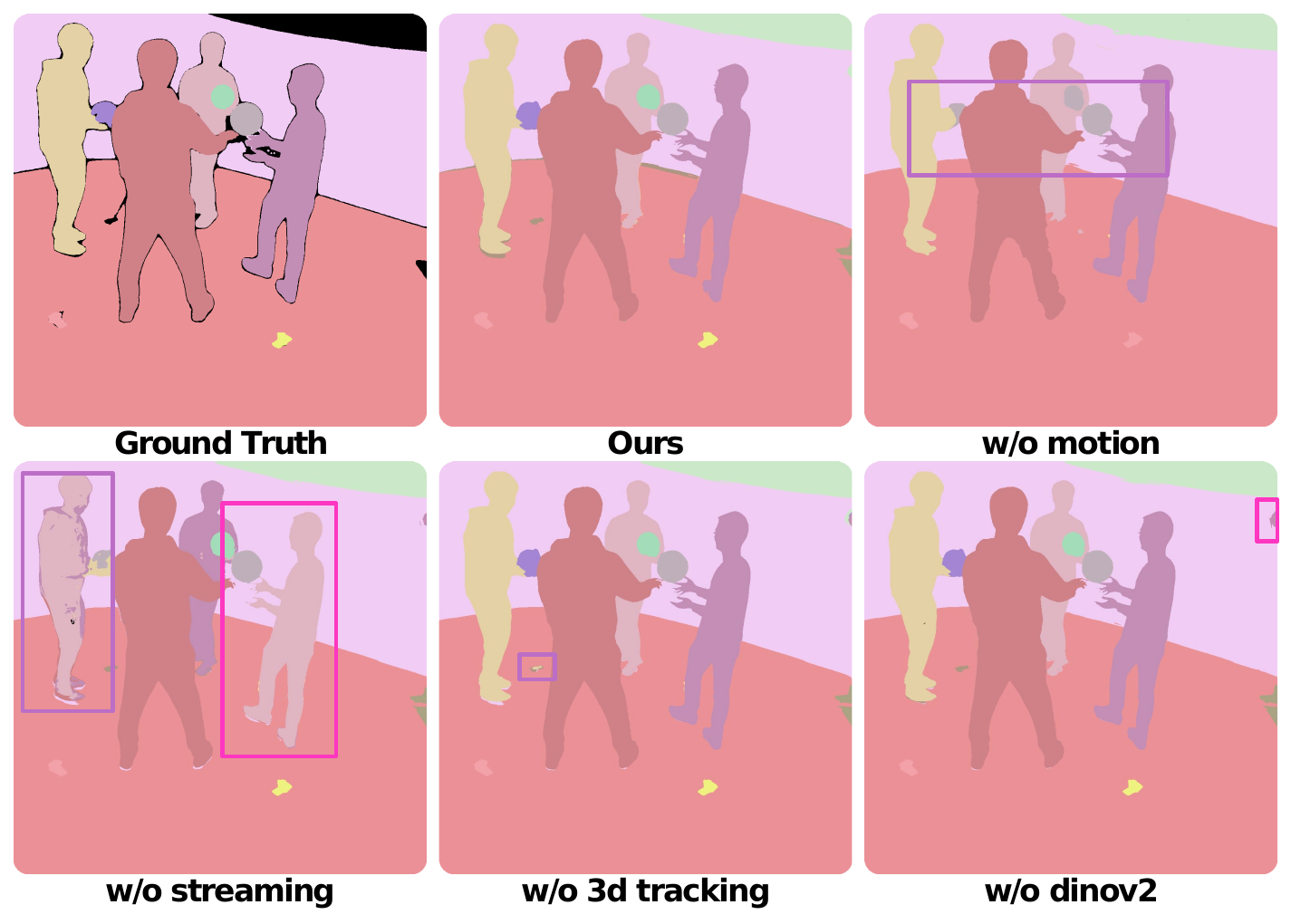}
    \caption{\textbf{Qualitative Results of Ablation Studies.}
    We present a qualitative comparison of segmentation results for different ablation settings on the Juggle sequence of the Multi-Human dataset.}
    \label{fig:ablation_result}
\end{figure}

\begin{figure}[bhtp]
    \centering
    \includegraphics[width=0.48\textwidth]{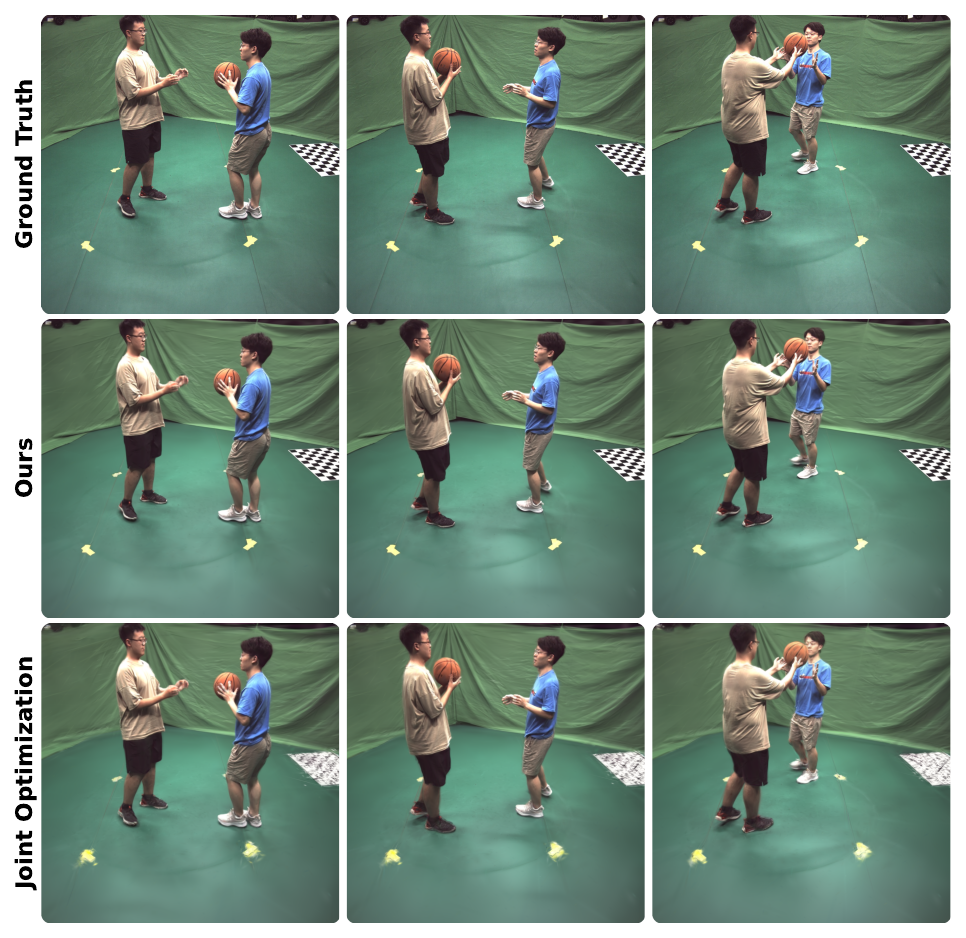}
    \caption{\textbf{Reconstruction quality comparison with joint optimization.} Joint optimization causes noticeable degradation in reconstruction quality due to multiple objectives.}
    \label{fig:ablation_learnable_gs}
\end{figure}

\subsection{Metrics and Baselines}

\paragraph{Metrics}  
We evaluate segmentation performance using four primary metrics: mean Intersection-over-Union (\(\text{mIoU}\)), mean accuracy (\(\text{mAcc}\)), recall, and F1 score.  
\(\text{mIoU}\) measures the average overlap between predicted and ground-truth masks across all selected instances, while \(\text{mAcc}\) computes average pixel-wise accuracy over the same set of instances.
Recall captures the proportion of correctly predicted foreground pixels, and F1 balances precision and recall to reflect overall segmentation quality.
To specifically assess performance on moving objects, 
we report a dynamic-region recall metrics \(\text{Recall}_{\text{dyn}}\), computed only over annotated dynamic instances.  
These provide a focused evaluation on challenging, motion regions that may be underrepresented in global averages.
\begin{figure*}[t!]
    \centering
     \includegraphics[width=0.95\textwidth]{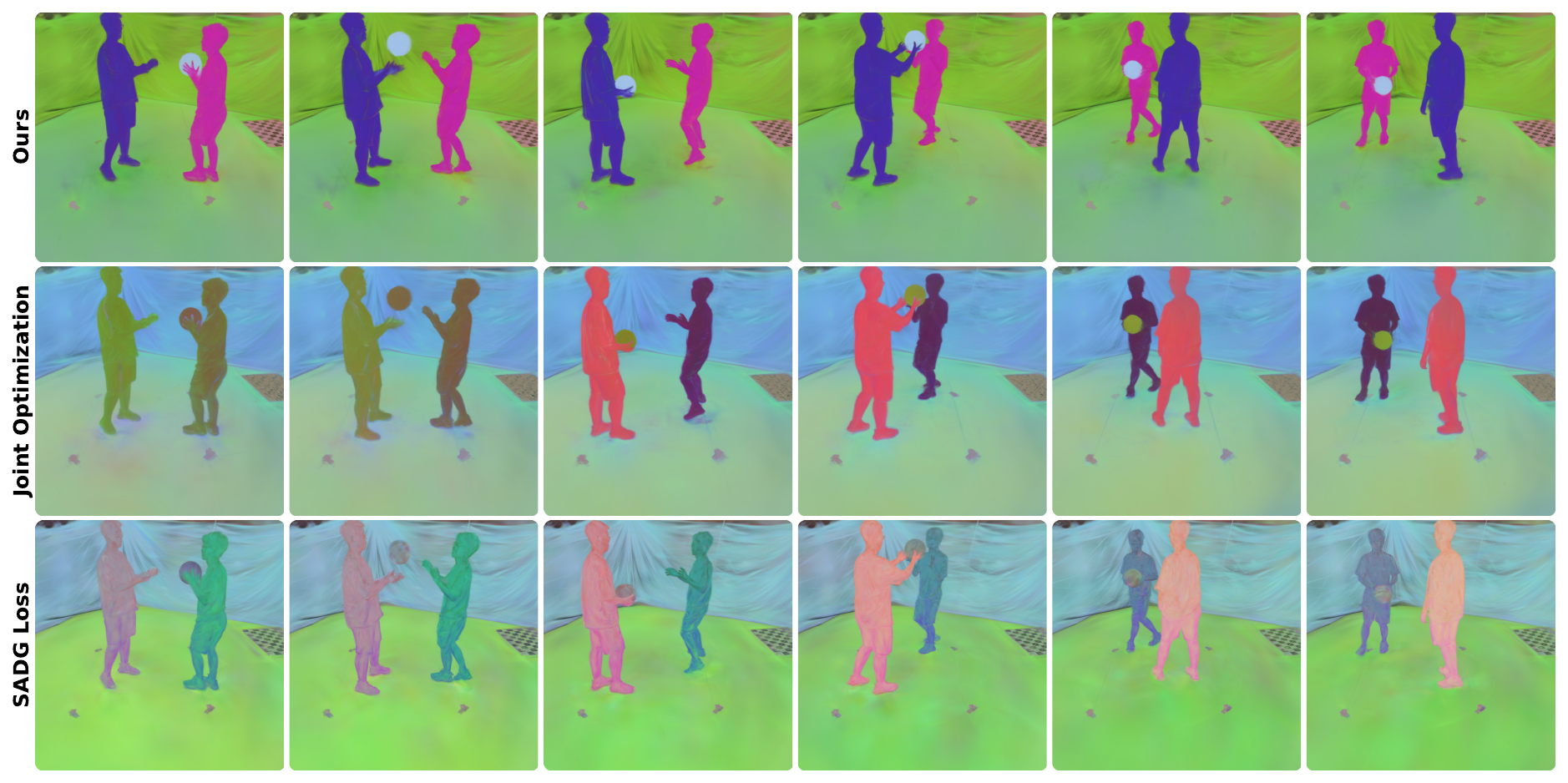}
    \caption{\textbf{PCA Visualization of Feature Learning.} Principal Component Analysis (PCA) of the learned feature fields under different feature learning strategies. Our method produces more temporally consistent and structured feature distributions across frames, indicating better alignment and robustness in dynamic scenes. This suggests that our feature learning framework effectively captures spatiotemporal coherence.}
    \label{fig:ablation_pca_result}
\end{figure*}

\paragraph{Baselines}
We compare our method with four baselines: OmniSeg3D~\cite{omniseg3d}, DGD~\cite{DGD}, SA4D~\cite{sa4d}, and SADG~\cite{sadg}.
OmniSeg3D, originally designed for static scenes, is integrated into the SADG backbone for fair comparison.  
DGD distills a feature field from DINOv2~\cite{Dinov2} features and is tailored for semantic segmentation; we manually annotate keypoints and apply a threshold to extract foreground masks.  
SA4D learns segmentation from DEVA~\cite{DEVA} predictions using the view closest to the test view.

\subsection{Comparison Experiments}

\paragraph{Neural3DV}

Qualitative and quantitative comparisons are presented in Fig.~\ref{fig:comparison_results} and Table~\ref{tab:comparison_result}, respectively.
As Table~\ref{tab:comparison_result} shows, our method attains the highest mIoU and mAcc on the Neural3DV dataset.
Although OmniSeg3D reports the best recall, it sacrifices precision, which leads to lower F1 and mIoU scores overall.
In dynamic scenes, our approach achieves a higher Recall$_\text{dyn}$, indicating superior handling of moving objects.
This advantage is visually evident in the first two rows of Fig.~\ref{fig:comparison_results}, where our segmentation remains accurate for both dynamic objects and static elements.

\paragraph{Multi-Human}
Qualitative and quantitative results are presented in Fig.~\ref{fig:comparison_results} and Table~\ref{tab:comparison_result}, respectively.
The Multi-Human dataset poses challenging situations with multiple people interacting and frequent human–object exchanges such as ball passing. 
As reported in Table~\ref{tab:comparison_result}, our method achieves consistently strong performance across all evaluation metrics.
SA4D exhibits high recall in dynamic regions, primarily because its predictions lack clear boundaries, resulting in oversized foreground regions that artificially inflate the metric.
Notably, DGD, which distills features fields from DINOv2, struggles in multi-person scenarios due to its tendency to merge different human instances into the same semantic category. 
This difference is clearly visible in the middle two rows of Fig.~\ref{fig:comparison_results}, where our approach maintains accurate segmentation despite large body motions and complex interactions that cause competing methods to fail.
Moreover, our method demonstrates unique capabilities in learning more precise segmentation results than the original supervision labels, 
even in cases where object boundaries are ambiguous. This enhanced performance primarily benefits from our Freetime Feature Learning, which leverages temporal motion information to optimize boundary localization. 
For example, as shown in Fig.~\ref{fig:ablation_invert_mask}, in human-object interaction scenarios such as ball passing, our method achieves more accurate boundary detection than SAM-generated segmentation results, particularly in challenging regions where hands interact with balls.

\paragraph{\textit{SelfCap}}
Qualitative and quantitative comparisons are presented in Fig.~\ref{fig:comparison_results} and Table~\ref{tab:comparison_result}, respectively.
We further evaluate our method on the challenging SelfCap dataset, which contains rapid motions and intricate human-object interactions.
As shown in the last two rows of Fig.~\ref{fig:comparison_results}, our approach reliably segments both human and pets, preserving accurate instance boundaries even on complex and high-speed action sequences.
\subsection{Ablation Study}

To validate the effectiveness of each component in our method, we conduct ablation studies on the Juggle and Basketball sequences from the Multi-Human dataset, particularly focusing on complex motion scenarios such as ball-passing sequences. 

\subsubsection{Ablation Studies on Model Components}

\begin{table*}[ht]
    \centering
    
    \caption{{\textbf{Component-wise Ablation Analysis on the Juggle Sequence of the Multi-Human Dataset.} We evaluate the impact of each architectural component by systematically removing it from our full model. The results demonstrate the importance of each component in maintaining segmentation accuracy across different metrics.}}
    \begin{tabular}{lcccccc}
    \toprule
    \textbf{} & \textbf{mIoU} & \textbf{mAcc} & \textbf{Recall} & \textbf{F1} & \textbf{\(\text{Recall}_{\text{dyn}}\)} \\
    \midrule
    w/o Motion & 0.644 & 0.995 & 0.713 & 0.686 &0.611 \\
    w/o Streaming Learning & 0.552 & 0.977 & 0.589 & 0.584 & 0.365 \\
    w/o 3D Tracking & \cellsecond 0.866 & \cellsecond 0.996 & 0.896 & \cellsecond 0.905 & \cellsecond 0.892\\
    w/o DINOv2 Alignment & 0.858 & \cellsecond 0.996 & \cellsecond 0.897 & 0.899 &  \cellsecond 0.892\\
    Ours & \cellfirst \textbf{0.895} & \cellfirst \textbf{0.997} & \cellfirst \textbf{0.928} & \cellfirst \textbf{0.934} & \cellfirst \textbf{0.898} \\
    \bottomrule
    \end{tabular}
    \label{tab:ablation}
\end{table*}

\begin{table}[ht]
    \centering  
    \caption{\textbf{Ablation study on regularization weights for the Juggle sequence.}
    We vary the 3D-tracking weight $\lambda_{1}$ and the DINOv2-alignment weight $\lambda_{2}$, and report mIoU, mAcc, Recall, $F_{1}$, and $\text{Recall}_{\text{dyn}}$.
    Bolded row indicates the configuration used in our final model.
   }
    \begin{tabular}{l|ccccc}
    \toprule
    $\lambda_1,\lambda_2$ & mIoU & mAcc& Recall & $F_1$ & \(\text{Recall}_{\text{dyn}}\) \\
    \midrule
    $10,10$ & 0.853 & 0.993 & 0.847 & 0.897 & 0.843\\
    $50,10$ & 0.863 & 0.997 & 0.890 & 0.903 &0.884 \\
    $100,5$ & 0.875 & 0.996 & 0.882& 0.901 & 0.865\\
    $\textbf{100},\textbf{10}$ & \textbf{0.895} & \textbf{0.997} & \textbf{0.928} & \textbf{0.934} & \textbf{0.898}\\
    $100,50$ & 0.854 & 0.997 & 0.887 & 0.894 & 0.875\\
    $100,100$ & 0.826 & 0.997 & 0.870 & 0.868 & 0.847\\
    $100,500$ & 0.591 & 0.957 &0.723 & 0.680 & 0.537\\
    $200,10$ & 0.892 & 0.997 & 0.917 & 0.931 & 0.881\\
    $500,10$ & 0.877 & 0.997 & 0.895 & 0.910 & 0.864\\
    \bottomrule
    \end{tabular}
    \label{tab:lambda}
\end{table}
\begin{figure}[ht]
    \centering
     \includegraphics[width=0.48\textwidth]{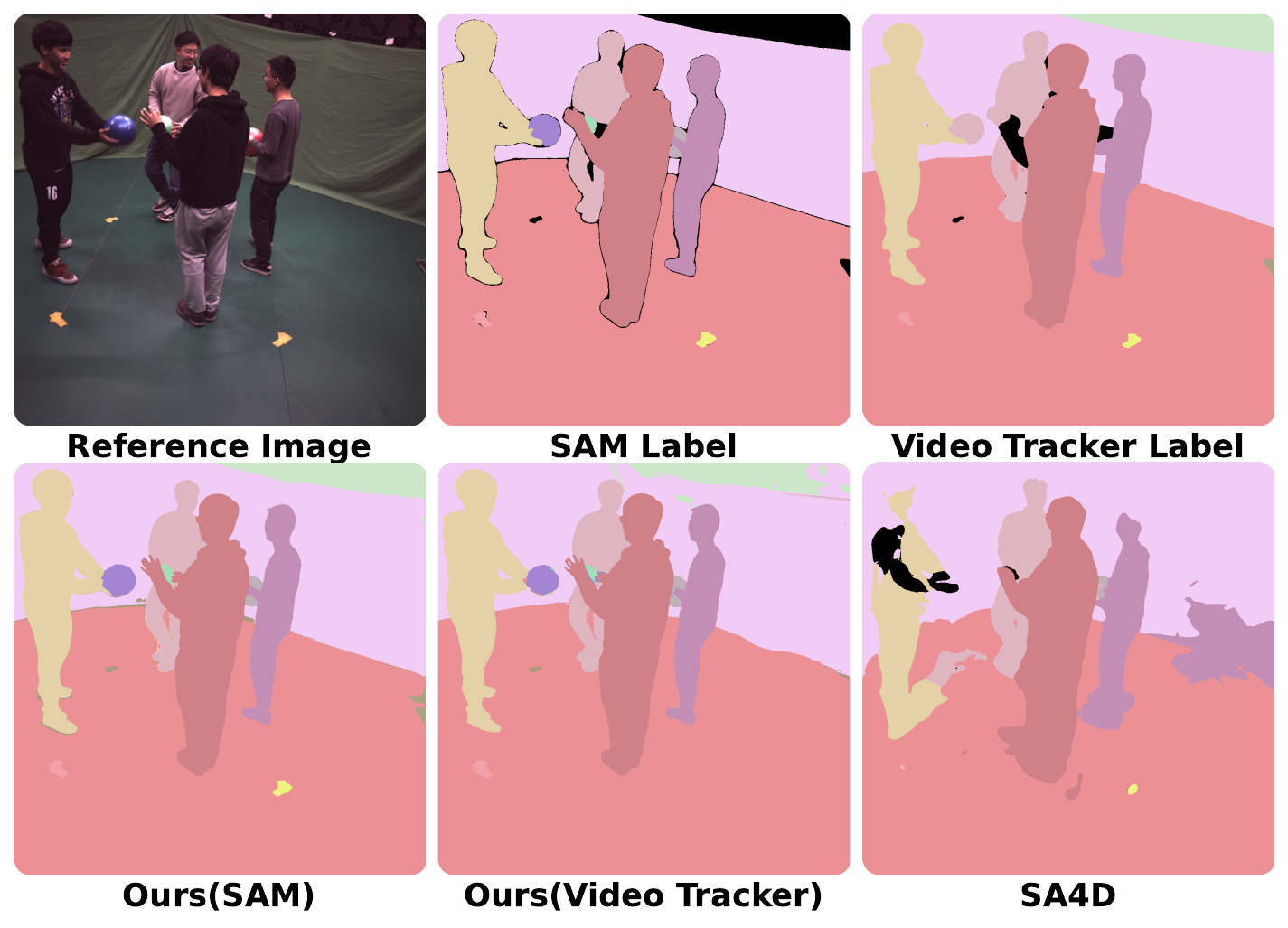}
    \caption{\textbf{Comparison with Video Tracker Supervision.} Performance evaluation of our method using labels generated by video trackers.}
    \label{fig:ablation_video_tracker}
\end{figure}

Qualitative and quantitative results for ablations on model components are shown in Fig.~\ref{fig:ablation_result} and Table~\ref{tab:ablation}.
\paragraph{Motion Modeling}
We analyze the importance of our explicit motion modeling by replacing it with the motion representation from 4DGS~\cite{fudan_4dgs}. 
As shown in Fig.~\ref{fig:ablation_result} and Table~\ref{tab:ablation}, removing our explicit motion modeling leads to significant performance degradation, particularly for objects with complex motions like balls in passing sequences. 
This degradation stems from the baseline 4DGS approach's coupling of motion with geometry optimization, which creates ambiguous velocity directions due to entangled spatial and temporal scales. 
Such ambiguity hinders the extension of local continuity to global scale, resulting in optimization challenges in the feature field for complex motion scenarios, as illustrated in Fig.~\ref{fig:ablation_streaming_and_motion}.

\paragraph{Streaming Sampling Strategy}
We evaluate the effectiveness of our streaming sampling strategy by replacing it with random sampling. 
As shown in Fig.~\ref{fig:ablation_result} and Table~\ref{tab:ablation}, random sampling leads to significant degradation in the segmentation performance, particularly for instances with large motions.
This degradation occurs because points of the same object at different timestamps tend to converge to different local optima, resulting in inconsistent feature representations across frames, as illustrated in Fig.~\ref{fig:ablation_streaming_and_motion}. 
To validate our streaming sampling strategy, we compare two single-frame optimization approaches: random camera sampling within a view and fixed-order camera sampling in basketball sequence. 
The two strategies achieve comparable results (mIoU: 0.9371 vs 0.9369, mAcc: 0.9973 vs 0.9968), confirming that the key to streaming sampling lies in feature propagation across frames rather than the specific sampling order within individual frames.

\paragraph{Additional Regularization}

We analyze the effectiveness of two additional regularization terms: 3D tracking and DINOv2 alignment. 
The 3D tracking regularization enforces spatiotemporal consistency between consecutive frames, while the DINOv2 alignment provides semantic guidance during feature learning. 
As shown in Table~\ref{tab:ablation}, both components provide complementary benefits, though their removal leads to only minor performance degradation.
To quantify their impact, we compute the mean intra-class feature variance (\(\overline{\sigma^2}\)) over time for dynamic object ground-truth masks. 
The variance decreases from \(\overline{\sigma^2}=\textbf{1.33}\) (w/o 3D tracking) to \(\overline{\sigma^2}=\textbf{1.28}\) (w/o DINOv2 alignment) to \(\overline{\sigma^2}=\textbf{1.12}\) (full model), demonstrating their contribution to temporal feature consistency.

We further analyze the impact of regularization weights \(\lambda_1\) (3D tracking) and \(\lambda_2\) (DINOv2 alignment) in Juggle sequence. 
As shown in Table~\ref{tab:lambda}, optimal performance is achieved with \(\lambda_1 = 100\) and \(\lambda_2 = 10\). 
While performance plateaus when \(\lambda_1\) exceeds 100, increasing \(\lambda_2\) beyond 500 leads to degradation due to interference with instance-level feature learning in multi-person scenarios. 
Moderate \(\lambda_2\) values provide beneficial semantic guidance in general scenes, while higher values can improve temporal feature alignment in scenes with significant semantic variations.

\subsubsection{Analysis of Freetime Feature Learning}
\begin{figure*}[t!]
    \centering
    
     \includegraphics[width=1.\textwidth]{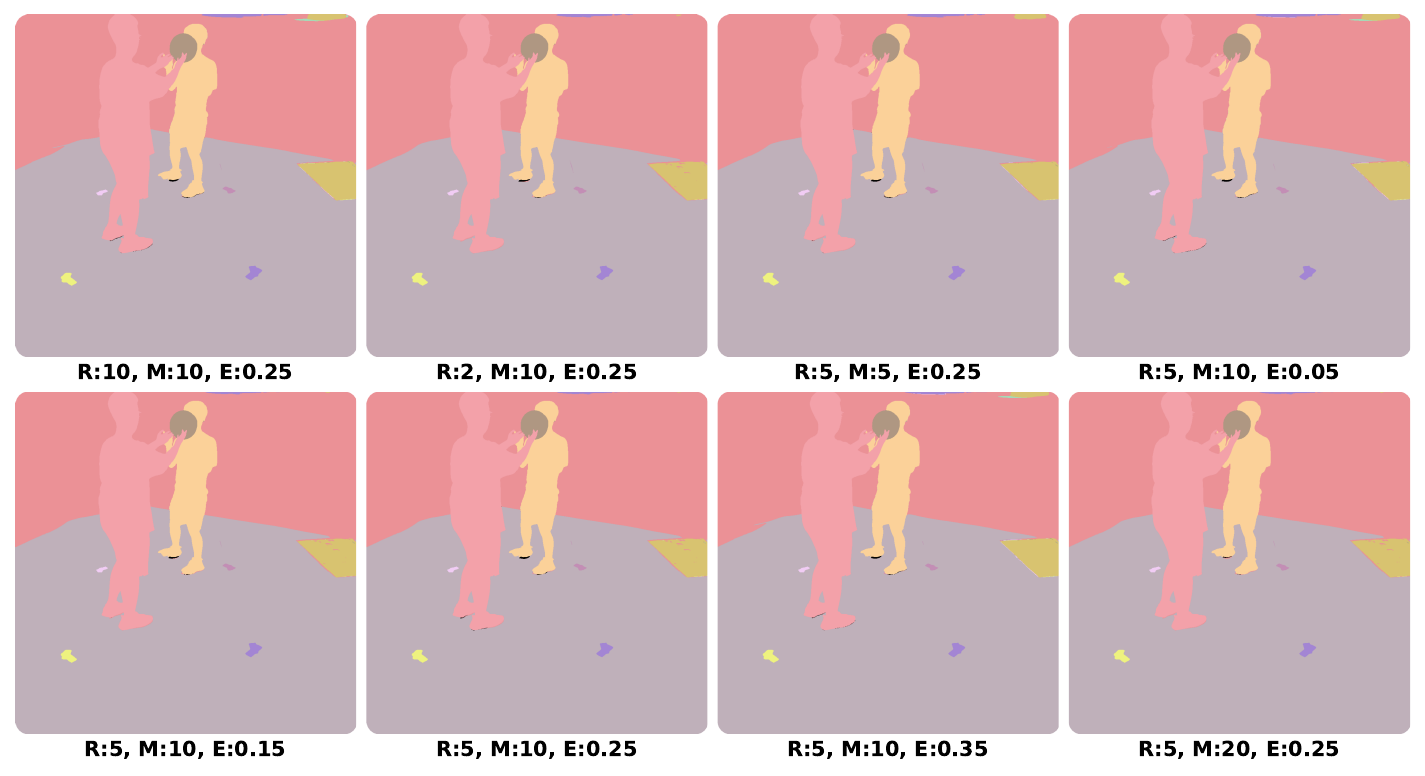}
    \caption{\textbf{Robustness Analysis of HDBSCAN Clustering.} We assess the robustness of our feature field segmentation with respect to key HDBSCAN hyperparameters, including the downsampling rate (\textbf{R}), the minimum number of samples (\textbf{M}), 
    and the minimum cluster selection epsilon (\textbf{E}). The results demonstrate that our method consistently achieves stable segmentation performance across a wide range of parameter settings, highlighting its strong robustness to clustering hyperparameters.}
    \label{fig:ablation_hdbscan}
\end{figure*}

\begin{figure*}[t!]
    \centering
     \includegraphics[width=1.\textwidth]{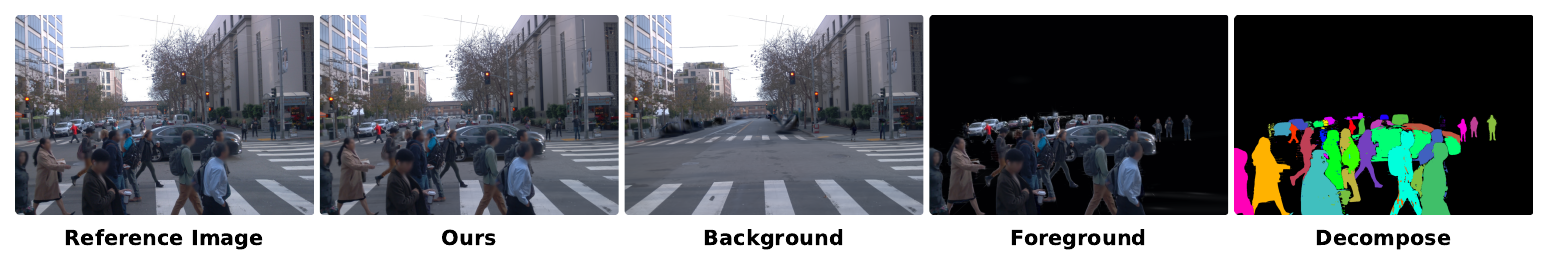}
    \caption{\textbf{Street Scene Decomposition in Extreme Scenarios.} We evaluate our method on a challenging Waymo street-scene clip featuring extreme crowding, heavy occlusion, and pronounced nonlinear motion. From left to right: reference image, our reconstruction result, background separation, foreground extraction, and instance decomposition. The results demonstrate our method's robustness in extreme dynamic scenarios.}
    \label{fig:challenge_result}
\end{figure*}

To validate the effectiveness and efficiency of our Freetime Feature Learning approach, we compare it with a baseline method built upon Freetime FeatureGS. 
The baseline performs per-frame feature learning using contrastive loss and propagates features through the Freetime FeatureGS motion model for 3D-to-4D segmentation. 
The baseline processes each frame in five main steps: (1) feature learning on the current frame using contrastive loss, (2) feature extraction and sampling of previous timestamp Gaussian primitives, (3) HDBSCAN clustering to obtain cluster labels, (4) label conflict resolution through local temporal consistency, where current labels are mapped to existing ones based on maximum probability matching and unmapped labels are treated as new parent labels, 
and (5) resetting global Gaussian primitive features and updating optimizer parameters for the next frame.

Although this baseline achieves reasonable segmentation quality, it has several limitations. 
First, its training speed is about 40 times slower than ours. 
Due to the requirement of sufficient feature learning in every frame and additional HDBSCAN clustering overhead. 
Moreover, segmentation errors in individual frames can propagate to subsequent frames, leading to error accumulation. 
In contrast, our method leverages temporal Gaussian reuse between consecutive frames through explicit motion modeling, 
which not only compresses the temporal representation but also accelerates feature learning by avoiding redundant optimization. 
This enables feature propagation across frames rather than isolated per-frame learning, resulting in more temporally consistent segmentation.

\begin{figure*}[h!]
    \centering
    \includegraphics[width=1.\textwidth]{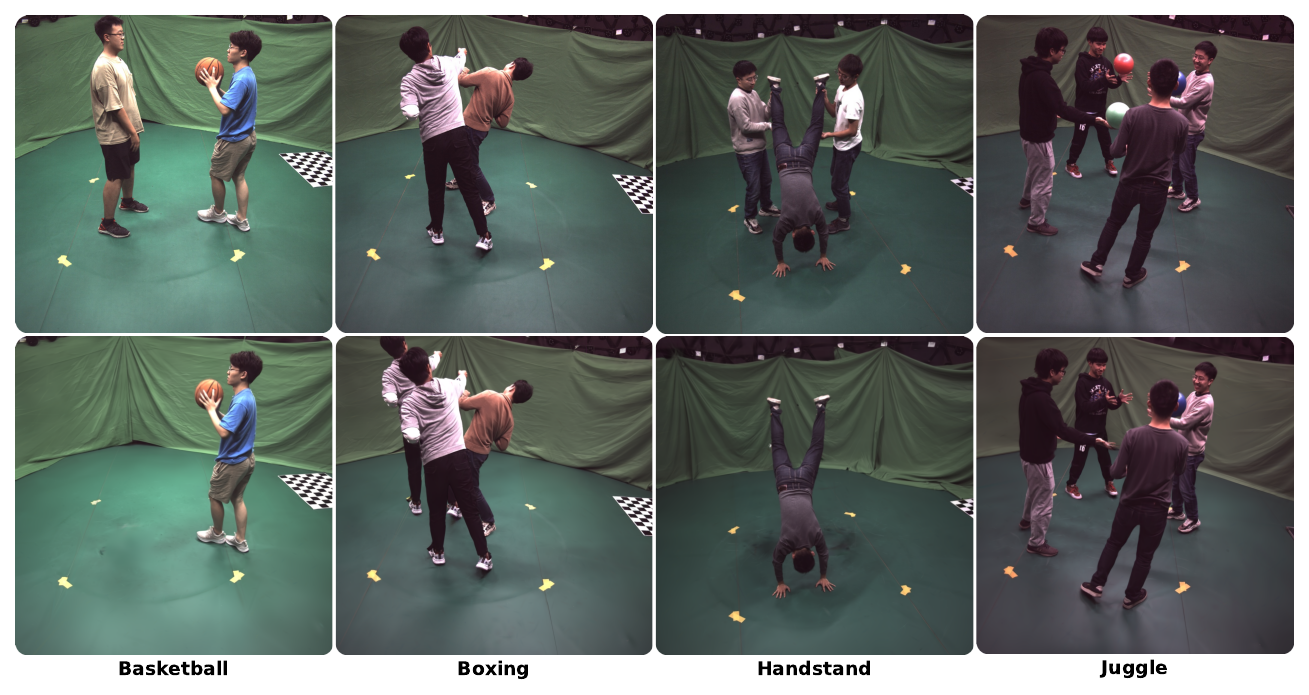}

    \caption{\textbf{Multi-Person Interaction Editing.} The top row displays the original scene, while the bottom row presents the edited results, including object/person duplication and removal.
    More results can be found in the supplementary video.
    }
    \label{fig:dynamic_scene_editing}
\end{figure*}

\begin{figure*}[h!]
    \centering

         \includegraphics[width=1.\textwidth]{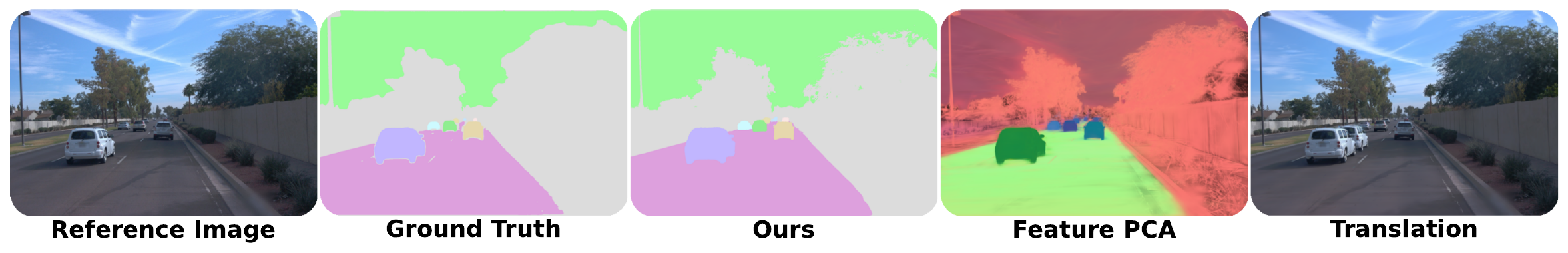}

    \caption{\textbf{Street Scene Decomposition.} We present the input RGB, ground-truth mask, predicted mask, feature map (PCA), and edited result.}
    \label{fig:monocular_street_scene}
\end{figure*}

\subsubsection{Analysis of Feature Learning Strategies}

We investigate the impact of various feature learning strategies on the Basketball sequence from the Multi-Human dataset.
Initially, we perform joint optimization on this sequence.
Following the initial reconstruction phase, we continue to optimize geometric and appearance-related Gaussian attributes using a reduced learning rate 5e-6, while introducing reconstruction features to examine feature learning behavior.
As illustrated in Fig.~\ref{fig:ablation_learnable_gs} and Fig.~\ref{fig:ablation_pca_result}, jointly optimizing for both reconstruction and segmentation objectives results in suboptimal performance for each task.
This degradation arises from conflicting optimization goals, which hinder the model’s ability to perform either task effectively.
Quantitatively, the PSNR of the reconstruction drops from 32.2dB (reconstruction-only) to 24.3dB (joint optimization), further confirming the adverse impact of task interference.

We further examine how different contrastive objectives influence feature learning.
The results in Fig.~\ref{fig:ablation_pca_result} show that compared with the contrastive objective from SADG~\cite{sadg}, our InfoNCE loss leads to more stable and less noisy feature representations.
This improvement can be attributed to our objective design, which provides stronger and more consistent supervision for learning discriminative features, enhancing the segmentation performance.

\subsubsection{Comparison of Supervision Strategies}

Although our method does not rely on video trackers, we further evaluate its performance when using labels generated by such less robust segmentation methods. 
Specifically, we adopt the same label acquisition approach as SA4D~\cite{sa4d} , utilizing video tracker outputs. Since our method does not require multi-view consistent supervision, we directly use the video tracker results from multiple training views as input. 
As shown in Fig.~\ref{fig:ablation_video_tracker}, our method maintains strong performance even when trained with these less reliable labels, despite the presence of significant errors in the video tracker results.
We acknowledge that the errors introduced by video trackers, compared to the segmentation results produced by SAM auto-masks, can negatively impact feature field learning. 
Nevertheless, our method exhibits remarkable robustness and consistently outperforms SA4D under these challenging conditions.

\subsubsection{Decoupling Analysis in Extreme Scenarios}
To further assess the generalization of our method in complex dynamic scenes, we evaluate it on a Waymo street-scene clip that is extremely crowded, heavily occluded, and exhibits pronounced nonlinear motion. The scene includes multiple motion patterns, such as pedestrians, cyclists, and vehicles.
As shown in Fig.~\ref{fig:challenge_result}, while maintaining clear foreground/background decoupling, our method achieves high completeness and boundary consistency for foreground instance segmentation, demonstrating strong competitiveness in extreme dynamic scenarios.
The robustness of our approach in such challenging conditions stems from the fundamental principle that any complex motion can be effectively approximated by a sequence of short linear motion segments, enabling our method to maintain temporal consistency even in the presence of highly nonlinear trajectories.

\subsubsection{Robustness Analysis of Clustering}

We further analyze the robustness of our feature field segmentation with respect to the HDBSCAN clustering process and the effectiveness of the subsequent filtering step. To evaluate the impact of clustering hyperparameters, we conduct experiments by varying the downsampling rate (\textbf{R}), the minimum number of samples (\textbf{M}), and the minimum cluster selection epsilon (\textbf{E}). 
As shown in Fig.~\ref{fig:ablation_hdbscan}, our method consistently achieves stable segmentation results across a wide range of parameter settings, demonstrating strong robustness to the choice of clustering hyperparameters.

%% file: sec/05_application.tex
\section{Applications}
\subsection{Dynamic Scene Editing}


We demonstrate the application of Dynamic Scene Editing on the Multi-Human dataset, as illustrated in Fig.~\ref{fig:dynamic_scene_editing}. 
Our method enables accurate and temporally consistent editing of dynamic scenes involving complex multi-person interactions. 
Specifically, it supports intuitive scene manipulations such as duplicating or removing specific individuals or objects across frames, which can be used for tasks like visual effects generation or behavior analysis.
Unlike prior approaches~\cite{ST-NeRF, multi_human}, which rely heavily on bounding box annotations as well as SMPL-based human priors and are typically limited to constrained scenarios, 
our method only requires multi-view 2D masks that are temporally independent. 
This greatly reduces the annotation burden and makes our pipeline more flexible and scalable.

\subsection{Monocular Street Scene Decomposition}%
We further validate our framework on monocular street scenes using the Waymo~\cite{waymo} dataset, as illustrated in Fig.~\ref{fig:monocular_street_scene}.
In this setting, our goal is to decompose complex urban environments, captured from a single camera view, into semantically meaningful regions that can support downstream tasks such as editing, scene understanding, and simulation.

Importantly, our method does not rely on expensive and labor-intensive 3D bounding box annotations~\cite{street_gaussians}, which are commonly required for monocular street scene reconstruction.
Instead, it operates using only RGB images and LiDAR depth for reconstruction, and leverages SAM to automatically generate 2D masks for training supervision.

%% file: sec/06_conclusion.tex
\section{Limitations}

\textbf{1)} \textit{Physical effects after Gaussian removal.} Current methods overlook 3D physical phenomena following the removal of Gaussian primitives (e.g., cast shadows), which may lead to artifacts such as visible holes or unmodeled occluded regions. Incorporating image inpainting models~\cite{lama} could plausibly fill in the missing content; \textbf{2)} \textit{Lighting consistency in editing/reconstruction.} During scene editing or reconstruction, directly inserting Gaussians fails to account for lighting consistency and other physical attributes. Employing relighting models~\cite{GSID, GIGS} or extending Gaussian representations to encode illumination-related properties~\cite{Gaussianshader} may yield more realistic and physically coherent results; \textbf{3)} \textit{Boundary quality under complex interactions.} While our method performs strongly in most scenarios, challenges remain for complex object interactions. The segmentation quality at dynamic object boundaries can be influenced by shared Gaussian primitives during reconstruction and by the precision of input 2D masks at object intersections, occasionally leading to less precise boundary delineation, especially under rapid motion or intricate interactions; \textbf{4)} \textit{3D tracking regularization under occlusions.} In extreme scenarios involving repeated or prolonged occlusions, explicit 3D correspondences may become unreliable, which in turn makes the neighborhood-search-based reliable matches in 3D tracking regularization sparse and noisy, degrading the constraint and potentially misguiding optimization.
Additionally, incorporating explicit inter-frame optical flow warping may further improve the robustness of the tracking regularization.~\cite{GauSTAR,RAFT}

\section{Conclusion}

We propose a novel framework for decomposed 4D scene representation and segmentation, built upon the Freetime FeatureGS representation. Our approach leverages the local continuity of this representation and extends it to a global feature field by introducing a streaming sampling strategy, 
which helps avoid local minima during feature field optimization. 
By combining the streaming sampling strategy with the motion relationships of Gaussian primitives in adjacent regions, our method employs a spatiotemporal contrastive loss to enable robust Freetime Feature Learning and achieve reliable feature field optimization.
Additionally, we introduce two regularization strategies to further enhance robustness, and propose an inference-time filtering strategy to provide better guidance for partially dynamic objects. Extensive experiments on multiple challenging datasets demonstrate that our method achieves superior performance compared to existing approaches, 
excelling in both segmentation accuracy and temporal consistency.

%% file: sec/07_acknowledgments.tex
\begin{acks}
The authors would like to acknowledge support from the National Key R\&D Program of China (No.~2024YFB2809102), NSFC (No.~62402427), ``Innovation Yongjiang 2035'' Key R\&D Program (No.~2025Z061), Ant Group, Information Technology Center, State Key Lab of CAD\&CG, and Zhejiang University Education Foundation Qizhen Scholar Foundation.
\end{acks}